\documentclass{article}
\usepackage{spconf,amsmath,graphicx,multirow,hyperref}

\makeatletter
\def\blfootnote{\xdef\@thefnmark{}\@footnotetext}
\makeatother

\title{FMG-Det: Foundation Model Guided Robust Object Detection}

\name{Darryl Hannan \qquad Timothy Doster \qquad Henry Kvinge \qquad Adam Attarian \qquad Yijing Watkins}

\address{Pacific Northwest National Laboratory \\
Seattle, WA}
\begin{document}

\maketitle

\begin{abstract}
Collecting high quality data for object detection tasks is challenging due to the inherent subjectivity in labeling the boundaries of an object. This makes it difficult to not only collect consistent annotations across a dataset but also to validate them, as no two annotators are likely to label the same object using the exact same coordinates. These challenges are further compounded when object boundaries are partially visible or blurred, which can be the case in many domains. Training on noisy annotations significantly degrades detector performance, rendering them unusable, particularly in few-shot settings, where just a few corrupted annotations can impact model performance. In this work, we propose FMG-Det, a simple, efficient methodology for training models with noisy annotations. More specifically, we propose combining a multiple instance learning (MIL) framework with a pre-processing pipeline that leverages powerful foundation models to correct labels prior to training. This pre-processing pipeline, along with slight modifications to the detector head, results in state-of-the-art performance across a number of datasets, for both standard and few-shot scenarios, while being much simpler and more efficient than other approaches.
\end{abstract}

\section{Introduction}
\label{sec:intro}
\blfootnote{© 2025 IEEE. Personal use of this material is permitted. Permission from IEEE must be obtained for all other uses, in any current or future media, including reprinting/republishing this material for advertising or promotional purposes, creating new collective works, for resale or redistribution to servers or lists, or reuse of any copyrighted component of this work in other works.}

\begin{figure}[t]
    \centering
    \includegraphics[width=0.65\columnwidth]{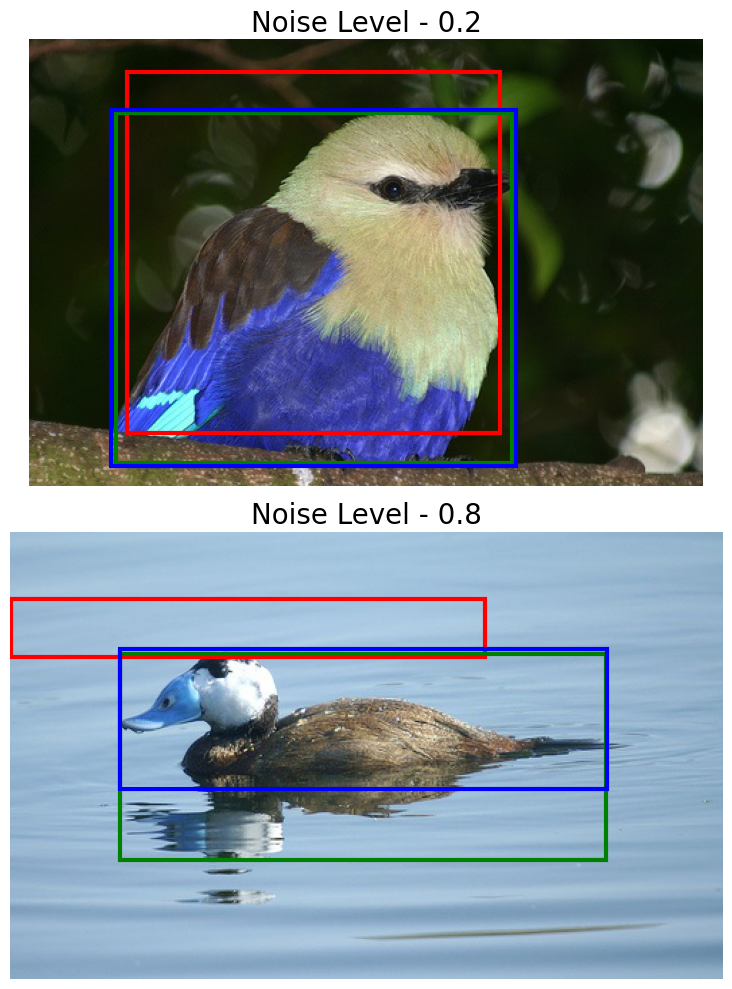}
    \caption{Even small amounts of noise in bounding box coordinates can have a significant impact on the parts of an object that are captured by the bounding box. This is visualized for two noise levels on VOC 2007 training examples (blue = original, red = noisy). In this paper, we propose a method to mitigate bounding box annotation noise (green = corrected by our proposed method, FMG-Det).}
    \label{fig:sample_noise}
\end{figure}

Collecting high quality annotations is challenging for object detection tasks; annotators must precisely and consistently place bounding boxes around the target objects. When objects sharply contrast with the background and have a relatively uniform shape, there is unlikely to be much variation in the specified locations. However, when objects are nonuniform, have poorly defined boundaries, or are occluded, the exact coordinates are not well defined, introducing substantial subjectivity and variance in labels across annotators. This problem is under-explored in object detection research, where highly curated popular datasets, such as MSCOCO \cite{lin2014microsoft}, are the standard by which object detection models are investigated.
When collecting data and training a detector for a given task, issues related to annotation quality are unavoidable, frequently difficult to mitigate, and can have a greater impact on performance than selecting an optimal architecture.

These problems have led to the development of robust object detection techniques, where the goal is to effectively train object detectors in noisy scenarios \cite{liu2022robust,ssddet,disco}.
However, prior work only explores and prioritizes preserving performance when the amount of noise is relatively minimal, demonstrating significant performance degradation as the amount of noise increases. This can be seen in Figure 1 in the supplementary, where noise is synthetically added to PASCAL VOC 2007 \cite{everingham2010pascal} on a scale from 0.0 (no noise) to 1.0 (significant noise) and a number of existing techniques are evaluated. At low noise levels, bounding box correction is hardly needed, as base model performance does not decrease significantly. However, as the amount of noise increases, the effectiveness of prior approaches is reduced.

Another problem that prior approaches fail to address is retaining performance in noisy limited data scenarios. While it might be possible to manually correct each instance in such cases, this is not always feasible. If the data is difficult to interpret and naturally lends itself to mislabeling, relabeling might prove ineffective. This restriction on the amount of training data adds an additional layer of complexity to robust object detection because the contribution of each individual example to model performance is much greater and, therefore, the impact of noise is greater as well.

In this work, we overcome both of these shortcomings by proposing a robust object detection algorithm that leverages the zero-shot capability of foundation models to automatically correct noisy bounding boxes. We call our method \emph{Foundation Model Guided Robust Object Detection (FMG-Det)}. At a high level, FMG-Det is a model-agnostic, zero-shot pre-processing pipeline that uses the Segment Anything Model (SAM) \cite{kirillov2023segany} and CLIP \cite{Radford2021LearningTV} to extract, score, and select a set of corrected bounding boxes from a set of noisy annotations.
We integrate this pre-processing pipeline into a multiple instance learning (MIL) framework, where we input both the corrected set of bounding boxes and the original noisy boxes to our detector and interpolate between the two sets of annotations using a parameterized module to select the optimal mixing ratio for each pair. Since the bulk of the computational overhead occurs during pre-training, this results in minimal additional compute to the model training process. Additionally, there are not many additional parameters that need to be learned, making the method well-suited for few-shot scenarios.

Our proposed methodology obtains an average performance gain of 6.8 mAP on COCO and 6.6 mAP on VOC compared to prior state-of-the-art approaches. We are also the first work to apply this same noise injection strategy to a few-shot detection task. We demonstrate that bounding box noise is more detrimental in few-shot scenarios and that our proposed architecture is once again most effective in mitigating the impact of this noise on detector performance.

\section{Related Work}
\label{sec:related_work}
Initial work in robust object detection focused on scenarios where both the bounding boxes and the class labels contained noise \cite{chadwick2019training,liu2022towards,li2020towards,mao2021noisy}.
Zhang et al. \cite{zhang2019learning} kicked off the line of work most similar to our own, exclusively focusing on bounding box noise. Liu et al. \cite{liu2022robust} took this work further, explicitly demonstrating that bounding box noise degrades overall performance more substantially than class label noise, while also focusing on more general object detection datasets. Bar et al. \cite{bar2023novel} introduced a benchmark dataset, based on PASCAL VOC, where noise was synthetically injected into the dataset. This has become the standard way of evaluating robust object detection models due to the difficulty in sourcing real world datasets that have noisy annotations for training and clean annotations for testing. Zhou et al. \cite{disco} introduced a technique that relied upon modeling the spatial distribution of object proposals, then based upon this distribution, correcting the class label or bounding box. Wu et al. \cite{ssddet} more directly built upon \cite{liu2022robust}, introducing spatial position and identity self-distillation modules, where the former improves the instance bags that are used for instance selection, while the latter predicts the IoU between a proposal and the object. Liu et al. \cite{liu2024dynamic} and Zhu et al. \cite{zhu2024robust} are also closely related works that explicitly focus on remote sensing.

Object-centric representation learning is a related field, where a model is trained to localize objects in a fully unsupervised setting  \cite{bao2022discovering,seitzer2023bridging,singh2022simple}. These models must also learn to handle noise in the form of object motion, as they are frequently trained on videos and leverage frame-to-frame changes to identify objects. Another line of work that is tangentially related to our own is that of open-vocabulary segmentation. Papers such as Grounded SAM \cite{ren2024grounded} also combine an object detector with Segment Anything \cite{kirillov2023segany}. This is similar to the first step of our proposed pipeline, where we use SAM to generate masks for noisy bounding boxes. However, this in and of itself is not a good solution to robust object detection due to the fact that this form of correction is fully dependent upon the quality of the SAM generations.
\begin{figure*}[t]
    \centering
    \includegraphics[width=0.9\textwidth]{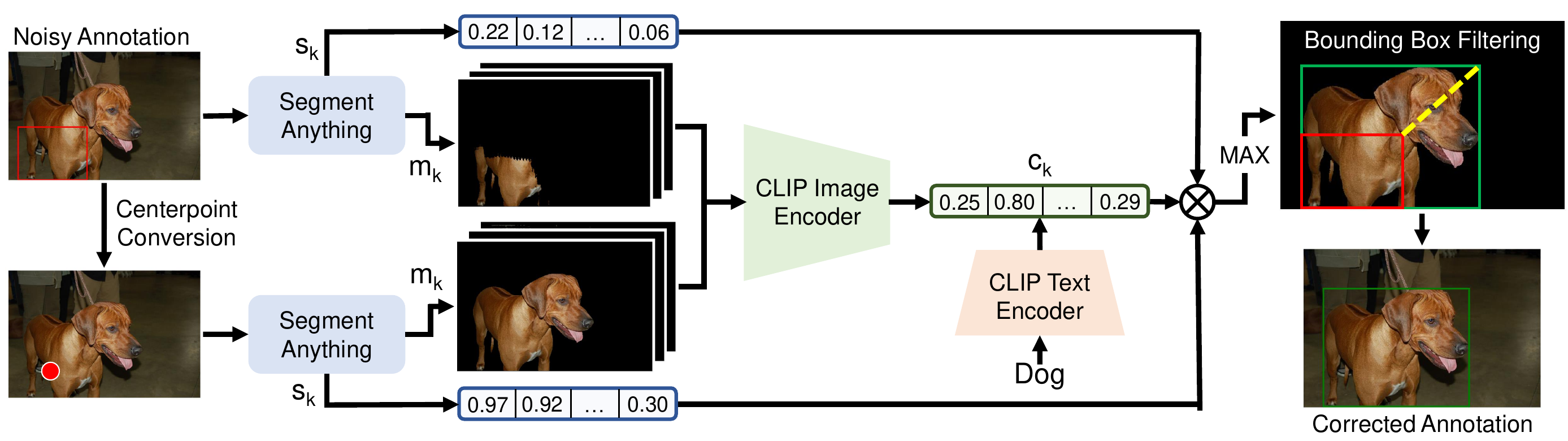}
    \caption{Overview of our proposed Foundation Model Correction (FMC) pre-processing pipeline. Segment Anything extracts a set of candidate regions and corresponding set of scores, using both point and bounding box prompts to diversify the set of masks that are produced, CLIP scores each region using the ground truth label, the CLIP and SAM scores are combined and the mask with the highest score is selected, and lastly, the mask is converted to a bounding box and it is compared against the original noisy annotation to ensure it did not shift too severely. This pipeline is run in a zero-shot fashion, completely offline.}
    \label{fig:fm_correction}
\end{figure*}
\section{Task Description} \label{sec:task_desc}
Consider a standard object detection task where a model is given an image $i$ and is tasked with producing a set of bounding boxes $B_i$ and corresponding set of labels $C_i$ for each target object in the image. Each bounding box $b_i$ consists of four coordinates $(x, y, w, h)$ representing the center coordinates and width/height of a box that fits tightly around the target object and each label $c_i$ is a single class selected from a fixed set of possible classes. In this work, we are primarily concerned with scenarios where the category labels are assumed to be accurate for all instances in the training set, but there is some noise applied to the bounding boxes $B_i$ such that they no longer fit tightly around the target object(s)\footnote{\cite{liu2022robust} previously demonstrated that bounding box noise has a much greater impact on detector performance than class noise.}. In general, we can define this noise as a function $f_N(B_i)$; the exact nature of this function will depend on the domain and will take the form of some standard set of mistakes or errors that humans tend to make.

Unfortunately, benchmarking robust object detection techniques is challenging. One would need a dataset that has noisy training data but clean test data for accurate evaluation. Therefore, it is common to synthetically generate noise \cite{liu2022robust, bar2023novel}. We follow \cite{liu2022robust}, where $f_N$ is defined independently for each bounding box coordinate. Values $\Delta_x$, $\Delta_y$, $\Delta_w$, and $\Delta_h$ are sampled from a uniform distribution on the interval $(-0.0, 1.0)$. This results in the bounding box being randomly shifted and scaled relative to its initial position. Some examples of this process can be seen in Figure \ref{fig:sample_noise}, where a value of 0.2 results in a slight shift in the bounding box but an example of 0.8 shifts the bounding box almost completely off the target object.
This task is more challenging than most real world scenarios because this random process is harder for a model to learn than consistent human errors.

\section{Method}
\subsection{Foundation Model Correction Pipeline}
We first propose a pre-processing methodology that directly adjusts the target bounding box to the target object, improving the quality of the training data by rectifying any potential errors that have been made and providing more consistent annotations to train on. We call this our Foundation Model Correction (FMC) pipeline. To accomplish this task, we use the generalist segmentation foundation model, Segment Anything \cite{kirillov2023segany} (SAM), and a vision-language model, CLIP \cite{Radford2021LearningTV}. An overview of this proposed pipeline is available in Figure \ref{fig:fm_correction}.

\subsubsection{Candidate mask generation.}
Consider an image $i$ and a set of noisy bounding boxes $B$. SAM takes an image as input, along with one or more prompts. These prompts allow the user to specify the objects in the image for which masks should be generated. Given that we already have $B$, we can use these to prompt the model. However, since the boxes are noisy, the prompts are suboptimal and can result in segmentation masks that do not properly cover the original target object. Therefore, we convert the bounding boxes to centerpoints $P$ and use these as an additional set of prompts that are separately passed to SAM. We then generate a set of masks $M$ and corresponding set of scores for those masks $S$ using two separate calls to SAM, i.e., $SAM(i, B)$ and $SAM(i, P)$. We concatenate these outputs together to get $M$ and $S$.

\subsubsection{Mask ranking and selection.}
We now must select the optimal mask from $M$. While using $S$ is viable, we propose improving the quality of these scores with CLIP \cite{Radford2021LearningTV}. To generate the CLIP scores, we extract each of the segmented regions from the original image, resulting in a set of images that only contain the masked areas. For a given ground truth bounding box $b_{j}$, where $j$ is the index of an object, we take the set of all predicted masks for that given location and pass these through the CLIP image encoder. We feed CLIP's text encoder a single phrase that corresponds to the groundtruth class at that given location, e.g., ``dog". We compute a score for each of the predicted masks via a softmax over the cosine similarities of each image encoding and the textual embedding representing the target class. This score is then combined with the score output by SAM and the maximum score is used to select the top scoring mask:
\begin{equation}
    \hat{m}_j = \max_{k}(\alpha * c_{j,k} + (1 - \alpha) * s_{j,k})
\end{equation}
where $\hat{m}_j$ is the top scoring mask for bounding box $b_{j}$, $c_{j,k}$ is the score output by CLIP for the $k^{th}$ mask for this object, $s_{j,k}$ is the score output by SAM for the same mask, and $\alpha$ is a hyperparameter that controls the contribution of the two scores. We then compute a new bounding box $\hat{b}_{j}$ by setting the new bounding box coordinates to the minimum and maximum $x$ and $y$ positions in $\hat{m}_j$. This results in a new set of `corrected' bounding boxes $\hat{B}$ for the given image that are tightly fit around the highest scored segmentation masks.

\subsubsection{Bounding box filtering.}
Our proposed procedure is contingent on SAM's capacity to segment the target object(s) accurately. While we found that SAM produces high quality results in most cases, there are certain scenarios in which SAM still fails to generate an accurate mask, either selecting a part of the object or a portion of the background in addition to the target object.
Therefore, we compute the Intersection over Union (IoU) between the noisy ground truth box and rectified box to capture both the change in position and change in size, and if the box shifted too substantially (by more than a fixed value $\lambda$), we keep the original noisy annotation.

\subsection{Foundation Model Guided Robust Object Detection}
\label{sec:FMG-Det}
After the FMC is run, we integrate the corrected boxes into a detection architecture.\footnote{We did explore directly integrating SAM into an OA-MIL style framework, where gradients flowed into SAM. However, we found that it was prohibitively expensive to train and did not produce significant improvements over the same model without direct SAM supervision.} This full pipeline is called \emph{Foundation Model Guided Robust Object Detection (FMG-Det)}. Since the FMC is detector agnostic, we opt to pair it with another denoising algorithm proposed by Liu et al. (2022) \cite{liu2022robust}, OA-MIL. This further improves instance refinement and selection over the course of training.

\subsubsection{Instance interpolation.}
We propose adding a new module to OA-MIL, an instance interpolation module.
We begin by passing both the initial set of noisy boxes $B$ into the network along with the corrected boxes $\hat{B}$, where $\hat{B}$ is the set of boxes directly output by the foundation model correction pipeline. Both of these sets have a 1-1 correspondence, where every noisy box $b_j$ has a corresponding corrected box $\hat{b_j}$. Given each pair of boxes, we train a network to output a mixing value $\gamma$ such that our interpolated bounding box is defined as:
\begin{equation}
    b^*_j = \gamma * \hat{b_j} + (1 - \gamma) * b_j
\end{equation}

To learn the appropriate value for $\gamma$ given the pair of bounding boxes, we train an additional network to predict this value given $\hat{b}_j$ and $b_j$. An illustration of this network is available in Figure 2 of the supplementary. Rather than relearning an entirely new network from scratch, we opt to repurpose the feature extraction capabilities of the object detector's backbone. We extract the region of interest (ROI) for both $\hat{b}_j$ and $b_j$ using the ROI extractor of the object detection network. $\gamma$ is computed as follows:
\begin{equation} \label{eq:interpolation}
    \gamma = \sigma(G_\theta( [ROI(\hat{b}_j, x); ROI(b_j, x)] ))
\end{equation}
where $G$ is a trained network with parameters $\theta$, $ROI()$ is the ROI align function, $x$ are the features from the backbone network for image $i$, $;$ is concatenation, and $\sigma$ is the sigmoid activation function. To maximize efficiency, $G$ is relatively small, consisting of just three feed-forward layers with ReLU activation functions. After instance interpolation, $b^*_j$ fully replaces $b_j$ as input to the instance generator and selector and the network is trained with the same objective function in \cite{liu2022robust}.
\section{Experiments}
\label{sec:experiments}
Following prior work \cite{liu2022robust,disco,ssddet}, we evaluate our model on MS-COCO \cite{lin2014microsoft} and PASCAL VOC \cite{everingham2010pascal} and synthetically add noise using the procedure outlined in Section \ref{sec:task_desc}. Since these splits are fully aligned with prior works, we do not exhaustively run all baseline models as all of the results that we present are directly comparable. We do rerun current state-of-the-art approaches, namely OA-MIL \cite{liu2022robust} and the current state-of-the-art technique, SSD-Det. We are the first work to present results at all noise levels from 0.0 to 1.0 at 0.2 intervals. We propose a new performance metric that summarizes the total performance of a given technique across all noise scenarios and use this as the primary metric. To calculate this metric we compute the mean absolute error (MAE) between the performance of the base detector with no noise and the final performance at each noise value across the 0.2 intervals.

\begin{table*}
    \centering
    \begin{tabular}{l l l l l l l || l}
    \multicolumn{8}{c}{PASCAL VOC}\\
    \hline
        Model & 0.0 & 0.2 & 0.4 & 0.6 & 0.8 & 1.0 & MAE \\\hline
        Faster RCNN & 77.3 ± 0.55 & 71.9 ± 0.25 & 44.3 ± 0.38 & 19.3 ± 0.35 & 13.5 ± 0.12 & 19.0 ± 0.52 & 36.4 \\
        OA-MIL \cite{liu2022robust} & 76.6 ± 0.12 & 73.4 ± 0.17 & 63.4 ± 0.17 & 35.5 ± 0.42 & 16.2 ± 0.25 & 18.2 ± 0.40 & 30.1 \\
        SSD-Det \cite{ssddet} & 77.9 ± 0.35 & 75.6 ± 0.31 & 67.5 ± 0.17 & 51.3 ± 0.38 & 32.5 ± 1.35 & 31.3 ± 1.46 & 21.3 \\\hline
        FMG-Det & 75.7 ± 0.36 & 73.2 ± 0.10 & 69.3 ± 0.32 & 62.6 ± 0.71 & 50.2 ± 0.10 & 46.5 ± 0.40 & \textbf{14.4} \\\hline
    \end{tabular}
    \caption{Mean average precision for PASCAL VOC dataset.}
    \label{tab:voc_results}
\end{table*}

\begin{table}[htbp] 
    \centering
    \begin{tabular}{l l l || l}
    \multicolumn{4}{c}{MSCOCO}\\
    \hline
        Model & 0.4 & 0.8 & MAE \\\hline
        Faster RCNN  & 8.5 & 0.7 & 33.1 \\
        OA-MIL \cite{liu2022robust} & 16.1 & 0.7 & 29.3 \\
        SSD-Det \cite{ssddet} & 27.1 & 1.5 & 23.4 \\\hline
        FMG-Det & 26.9 & 15.7 & \textbf{16.4} \\\hline
    \end{tabular}
    \caption{Mean average precision for MSCOCO dataset.}
    \label{tab:mscoco_results}
\end{table}

We also consider each of these datasets in few-shot scenarios.
We adopt the same few-shot data splits that were created in \cite{wang2020frustratingly} for PASCAL VOC, where classes were split into sets of base and novel classes, the former being for initial training and the latter being for fine-tuning. We use the first 10 seeds from the first base/novel split and the same synthetic noise procedure outlined above to inject noise into both the base and the novel splits.

Hyperparameters, other experimental details, limitations, and ablations are available in the supplementary.

\subsection{Results}
\subsubsection{Full Training}
Table \ref{tab:voc_results} contains our primary results for VOC PASCAL \cite{everingham2010pascal}.
The upper performance bound for Faster RCNN, the base detector for all models in this work, is 77.3 on VOC; the goal is then to retain this performance as the amount of noise increases. According to the MAE, it is clear that FMG-Det beats prior state-of-the-art approaches by a significant margin, with it being a full 6.6 points lower than the previous best method, SSD-Det. In the more fine-grained breakdown of performance across each noise level, we can see that at a noise level of 0.0 or 0.2, our method underperforms prior approaches by a couple of points. However, it is in these noise settings that denoising is least critical.
At a noise level of 0.4 and above, the gap in performance between our approach and all other approaches widens, resulting in up to a 16 mAP improvement over SSD-Det at a noise level of 1.0.

Table \ref{tab:mscoco_results} contains results for COCO \cite{lin2014microsoft}. Due to the extensive time required in training a model on COCO, we only evaluate the performance of the detector at two noise levels.
Our model substantially outperforms the prior state-of-the-art approach in MAE by 6.8 points. Taking a more fine-grained look at the performance on each individual noise level, we can see that at a noise level of 0.4, our approach performs virtually on par with the prior state-of-the-art technique. However, at a noise level of 0.8, all prior techniques fail to train a model that is even capable of performing the task, obtaining an mAP of less than 2. COCO is a more challenging dataset than PASCAL VOC; while prior approaches were able to achieve some results at high noise levels on the latter dataset, the challenge of the former, combined with the substantial noise, makes the task virtually impossible for them to learn.

\begin{figure}
    \centering
    \includegraphics[width=0.95\columnwidth]{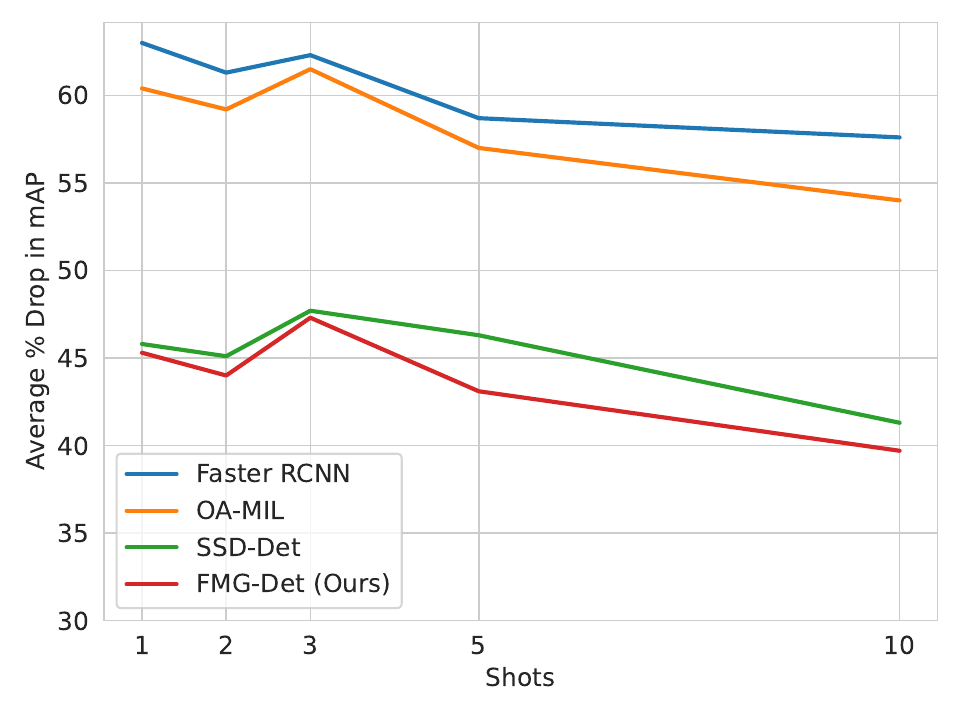}
    \caption{Illustration of how model performance is impacted as noise increases in few-shot settings for the PASCAL VOC dataset. The average drop in mAP is calculated by taking the MAE of the model across all noise levels as a percentage of the base model's performance with no noise.}
    \label{fig:few_shot}
\end{figure}

\subsubsection{Few-shot Detection}
Our primary results for few-shot detection on the VOC PASCAL dataset are available in Figure \ref{fig:few_shot} with a more detailed performance breakdown in Table 1 in the supplementary. We include results using just our FMC pipeline with Faster RCNN as an additional baseline and see a substantial improvement in performance beyond OA-MIL, however, our full FMG-Det architecture improves performance further, beating SSD-Det. We notice that in very limited data settings, the gap in performance between FMG-Det and SSD-Det is not as large when compared to 5+ shots or the full PASCAL VOC and COCO datasets. We hypothesize that this is due to the fact that slight errors created by FMC can have a substantial impact on performance in very limited settings, despite our best efforts to equip FMG-Det with filters and learned modules to mitigate these errors.

We generally see that in the few-shot setting, bounding box noise is even more detrimental, where adding noise during both base pre-training and novel fine-tuning in the most extreme few-shot settings quickly results in a final model that obtains close to 0 mAP. At very low shots (1-3) there does not seem to be a clear pattern. However, as the number of shots increases, there is a relative drop in performance degradation, suggesting that both the base model and the bounding box correction techniques are less effective in few-shot scenarios compared to standard data settings.

\section{Conclusion}
Object detector performance degrades significantly if bounding boxes are not carefully placed over the target objects.
We presented a new robust object detection algorithm, FMG-Det, that leverages powerful foundation models to correct bounding boxes, mitigating the impact of noisy annotations. We demonstrated through rigorous experimentation on standard datasets that FMG-Det significantly improves detector performance, especially in high noise scenarios, beating all prior state-of-the-art approaches. We similarly showed that our approach outperforms all prior work in few-shot object detection scenarios, which have not been investigated previously in the context of robust object detection. We hope that this work will support new object detection capabilities for downstream tasks that have limited, potentially noisy training datasets.

\bibliographystyle{icip}
\bibliography{icip}

\clearpage
\appendix

\section{Effectiveness of Prior Approaches in High Noise Scenarios}

\begin{figure}
    \centering
    \includegraphics[width=\columnwidth]{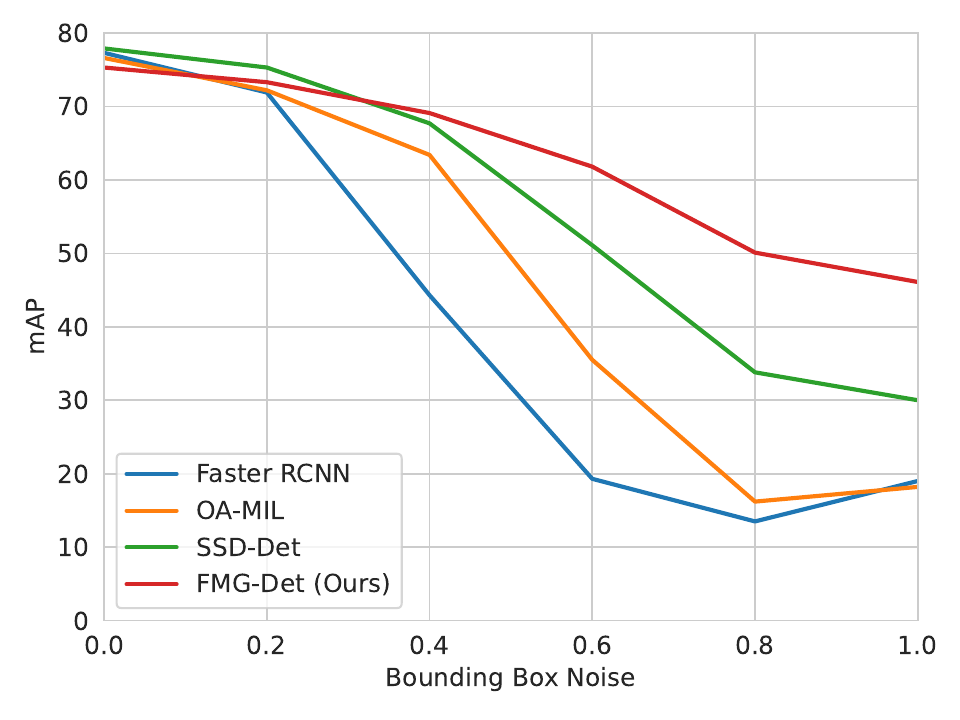}
    \caption{VOC test mAP demonstrating the considerable impact of bounding box noise on model performance in prior state-of-the-art models, including OA-MIL and SSD-Det, compared to our proposed FMG-Det algorithm.
    }
    \label{fig:pascal_map}
\end{figure}

Figure \ref{fig:pascal_map} presents results on PASCAL VOC at various noise levels, ranging from 0.0 (no noise) to 1.0 (severe noise). Prior work only focused on noise levels up to 0.4; opting to not even run their models at higher noise levels. The Faster RCNN model is just a standard detector without any noise mitigation approaches. OA-MIL \cite{liu2022robust} and SSD-Det \cite{ssddet} are prior noise mitigation approaches that are added to a Faster RCNN model. As the amount of noise increases beyond 0.4, the mAP deteriorates significantly for each of these approaches. Our proposed method, FMG-DET, addresses this shortcoming, retaining strong performance even under severe noise.

\section{Instance Interpolation Module}
\begin{figure}
    \centering
    \includegraphics[width=0.9\columnwidth]{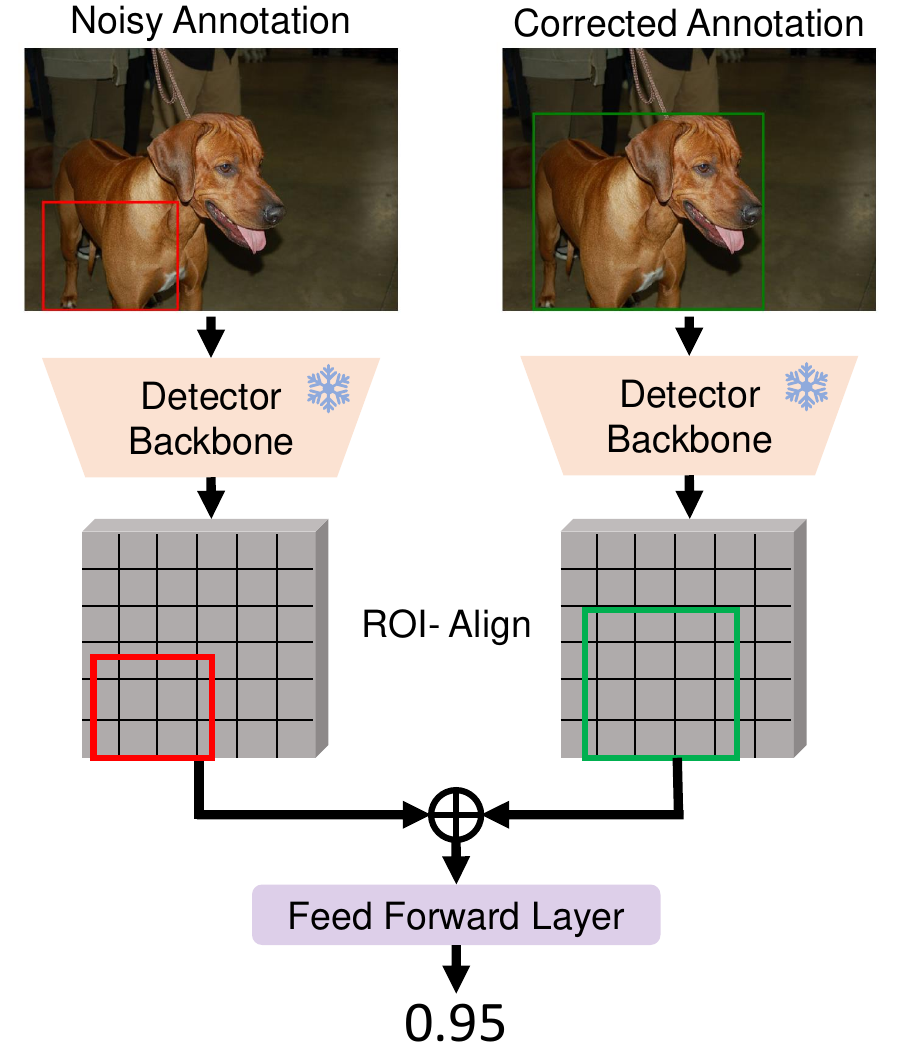}
    \caption{An overview of our proposed Instance Interpolation module. Both the corrected and noisy bounding boxes are passed to this module. It then extracts features for each bounding box using the backbone that already exists in the detector, and using these features, predicts a value $\gamma$ that is used to then interpolate between the corrected and noisy boxes.}
    \label{fig:interpolation_figure}
\end{figure}
Figure \ref{fig:interpolation_figure} provides an overview of our Instance Interpolation module. The details of this module are available in the main paper.

\begin{table*}[t]
    \centering
    \begin{tabular}{l l l l l l l l l}
    \hline
        Shots & Model & 0.0 & 0.2 & 0.4 & 0.6 & 0.8 & 1.0 & MAE \\\hline
        \multirow{3}{*}{1-shot} & Faster R-CNN & 19.2 & 15.3 & 5.1 & 0.8 & 0.4 & 1.7 & 12.1 \\
        & OA-MIL & 22.1 & 14.7 & 4.9 & 1.4 & 0.7 & 2.1 & 11.6 \\
        & FMC & 17.6 & 15.3 & 7.8 & 3.5 & 2.6 & 1.3 & 11.2 \\
        & SSD-Det & 21.8 & 18.3 & 14.7 & 4.7 & 1.1 & 1.6 & 8.8 \\\hline
        & FMG-Det & 21.4 & 19.2 & 10.8 & 6.0 & 3.3 & 2.4 & \textbf{8.7} \\\hline\hline
        \multirow{3}{*}{2-shot} & Faster R-CNN & 37.7 & 28.7 & 12.0 & 3.6 & 2.1 & 3.3 & 23.1 \\
        & OA-MIL & 41.2 & 31.5 & 11.5 & 3.6 & 1.6 & 3.1 & 22.3 \\
        & FMC & 36.5 & 29.0 & 17.9 & 11.1 & 8.0 & 5.5 & 19.7 \\
        & SSD-Det & 40.1 & 33.8 & 28.4 & 11.9 & 5.4 & 4.4 & 17.0 \\\hline
        & FMG-Det & 40.4 & 35.7 & 22.9 & 13.7 & 7.4 & 6.4 & \textbf{16.6} \\\hline\hline
        \multirow{3}{*}{3-shot} & Faster R-CNN & 52.2 & 37.8 & 14.3 & 5.4 & 3.3 & 5.2 & 32.5 \\
        & OA-MIL & 52.3 & 39.1 & 15.8 & 5.6 & 3.3 & 4.3 & 32.1 \\
        & FMC & 50.1 & 40.7 & 25.6 & 15.3 & 12.2 & 8.8 & 26.8 \\
        & SSD-Det & 51.9 & 44.8 & 37.2 & 15.9 & 7.4 & 6.5 & 24.9 \\\hline
        & FMG-Det & 51.6 & 41.9 & 30.9 & 17.9 & 11.2 & 11.1 & \textbf{24.7} \\\hline\hline
        \multirow{3}{*}{5-shot} & Faster R-CNN & 59.6 & 44.7 & 21.2 & 8.0 & 5.6 & 8.3 & 35.0 \\
        & OA-MIL & 61.1 & 49.4 & 23.4 & 8.3 & 5.3 & 6.3 & 34.0 \\
        & FMC & 59.6 & 49.0 & 33.4 & 20.9 & 17.2 & 14.2 & 27.2 \\
        & SSD-Det & 58.1 & 49.7 & 42.8 & 21.8 & 10.7 & 8.7 & 27.6 \\\hline
        & FMG-Det & 59.7 & 50.6 & 39.5 & 23.3 & 16.3 & 14.0 & \textbf{25.7} \\\hline\hline
        \multirow{3}{*}{10-shot} & Faster R-CNN & 63.0 & 49.5 & 22.2 & 8.9 & 6.2 & 10.3 & 36.3 \\
        & OA-MIL & 63.6 & 55.6 & 28.7 & 11.3 & 5.9 & 7.0 & 34.3 \\
        & FMC & 62.2 & 52.4 & 40.5 & 24.6 & 20.3 & 18.2 & 26.6 \\
        & SSD-Det & 62.4 & 56.1 & 50.1 & 26.5 & 13.2 & 13.6 & 26.0 \\\hline
        & FMG-Det & 61.8 & 57.4 & 44.5 & 26.9 & 20.0 & 17.5 & \textbf{25.0} \\\hline
    \end{tabular}
    \caption{Mean average precision for few-shot PASCAL VOC Novel Set 1 dataset.}
    \label{tab:voc_few-shot_results}
\end{table*}

\begin{table*}[t]
    \centering
    \begin{tabular}{l l l l l l l || l}
    \hline
        Model & 0.0 & 0.2 & 0.4 & 0.6 & 0.8 & 1.0 & MAE \\\hline
        Faster RCNN & 77.3 & 71.9 & 44.3 & 19.3 & 13.5 & 19.0 & 36.4 \\
        FM Correction & 75.0 & 72.1 & 66.1 & 55.2 & 46.5 & 44.0 & 17.5 \\
        FM Correction + OA-MIL & 75.1 & 72.2 & 67.0 & 55.5 & 44.4 & 41.4 & 18.0 \\
        FM Correction + Instance Interpolation & 76.6 & 73.4 & 67.8 & 58.2 & 48.7 & 44.9 & 15.7 \\\hline
        FMG-Det & 75.7 & 73.2 & 69.3 & 62.6 & 50.2 & 46.5 & \textbf{14.4} \\\hline
    \end{tabular}
    \caption{Ablations for FMG-Det using the Pascal VOC 2007 dataset, demonstrating the performance impact of each component. Starting from Faster RCNN, FMG-Det adds the Foundation Model Correction (FM Correction) pipeline, instance interpolation, and then leverages OA-MIL.}
    \label{tab:ablations}
\end{table*}

\section{Full Few-shot Detection Results}
Table \ref{tab:voc_few-shot_results} contains the full results for our few-shot experiments in the main paper, highlighting the performance of each approach at all noise levels for each few-shot scenario.

\section{Computational Efficiency of FMC Pipeline}
While the foundation model correction pipeline involves large foundation models, it can be run entirely offline with the remainder of the FMG-Det architecture being leveraged to prioritize performance, or a more efficient detector leveraged to prioritize training and inference time. For the experiments in this paper, due to compute limitations, we ran our experiments with a batch size of 1 on a single Tesla V100. Computation scales primarily with the number of images but also with the number of bounding boxes in each image. For COCO, this resulted in a rate of 2.475 seconds/image, and for VOC, a rate of 1.350 seconds/image. Note that due to the pipeline being training-free, it is highly conducive to distribution across multiple GPUs, where the dataset can easily be sharded, dramatically increasing the inference speed.

\section{Full Ablations}

Table \ref{tab:ablations} contains ablations for our proposed FMG-Det model, starting from the base detector, Faster RCNN \cite{ren2015faster}, and adding each of our proposed components, along with OA-MIL. Our Foundation Model Correction pipeline makes the largest contribution to the overall performance of our proposed approach, improving MAE from 36.4 to just 17.5. Relative to state-of-the-art approaches, just including this pipeline already achieves state-of-the-art performance on PASCAL VOC. This is critical as it is fully detector agnostic and therefore it is reasonably assumed that virtually any object detector would enjoy similar benefits. Interestingly, adding OA-MIL directly on top of this pipeline slightly decreases performance. However, adding our instance interpolation module to OA-MIL does boost performance further by a substantial margin. We did explore using SSD-Det instead of OA-MIL to see whether performance could be improved further by simply swapping them. Yet, we found that this, similar to adding OA-MIL directly on top of the foundation model correction pipeline, did not result in further performance improvements, suggesting that there is some redundancy between the learned denoising procedure in SSD-Det and our own proposed contributions. We hypothesize that adding the FMC pipeline diminishes the issues of object drift and group prediction, two of the primary issues with OA-MIL that motivated SSD-Det, reducing the effectiveness of the latter's improvements.

\subsection{Experimental Details}
Our model was built off of the OA-MIL \cite{liu2022robust} repository, which is in turn built on top of MMDetection \cite{mmdetection}. We use Faster R-CNN \cite{ren2015faster} with a ResNet-50 \cite{he2016deep} backbone as our object detector architecture due to its simplicity and the fact that we are not prioritizing overall model performance, rather we are focused on improving model robustness. However, note that FMG-Det is directly compatible with any 2-stage detector and the Foundation Model Correction pipeline is compatible with virtually any detector or alternative denoising technique. We use many of the defaults provided by MMDetection for the Faster R-CNN model. For our experiments on the full VOC and COCO datasets, we use a batch size of 8 for all of our experiments. All models are trained using the standard 1x learning schedule, which is run over 12 epochs and consists of SGD with a learning rate of 0.02, momentum of 0.9, a weight decay of 0.0001, a warmup linear scheduler that executes over 500 iterations with a warmup ratio of 0.001, and a multistep scheduler with milestones at 8 and 11 epochs with a gamma value of 0.1. For the OA-MIL \cite{liu2022robust} and SSD-Det \cite{ssddet} baselines, we used the defaults provided in the authors' repositories. For our foundation model correction pipeline, we used an $\alpha$ of 0.5 to mix the scores from SAM and CLIP and a $\lambda$ of 0.05 as our IoU threshold for accepting a correction. We selected these values by empirically running on subsets of the data and adapting them to minimize the number of dramatically shifted bounding boxes, e.g., egregious mistakes such as placing the bounding box around the background rather than the target.

We use the same training procedure outlined in \cite{wang2020frustratingly} to train our few-shot models, where a base model is trained, the weights are frozen except the head, then the model is finetuned on a mixed base+novel few-shot set. We train the base model using each of our proposed techniques and baselines, using the SAM hyperparameters that are outlined above. We do leverage slightly different hyperparameters for fine-tuning on the novel set. Namely, we fine-tune most of the models with a learning rate of 0.1 and a fixed learning schedule. The only exception is SSD-Det \cite{ssddet}, which we found works best with the same learning scheduler as above, just with a lower learning rate of 0.01.

\section{Limitations}
Our proposed approach is reliant upon the performance of SAM on the target dataset. While SAM has demonstrated exceptionally strong performance across a wide variety of domains \cite{kirillov2023segany}, in highly specialized domains that involve distinct image modalities, such as medical imaging, our proposed approach might not be effective if the quality of the extracted masks is low. However, this issue can be mitigated by leveraging a SAM variant that is better suited for the target dataset, whether it is a model like MedSAM \cite{MedSAM}, which has been specifically trained for medical imagery, or simply a SAM model that has been finetuned on images that are more closely aligned with the detection task. Another limitation of our proposed approach is that it struggles with bounding boxes that have no overlap with the groundtruth object. In these cases, SAM is likely to segment part of the background or an adjacent object. This can result in severe failure scenarios, where the box becomes more inaccurate than the noisy groundtruth. We attempt to mitigate such scenarios by discarding corrected boxes with no overlap, defaulting to the noisy groundtruth. However, this solution negates the benefits of the foundation model correction pre-processing step. Lastly, we were also unable to test our approach on noisy, publicly available datasets that could be reported in our paper. This dataset would need to have naturally noisy training labels but clean, high-quality testing labels for validation. However, as discussed in the main paper, we do believe that our synthetic setting is more challenging than most real world settings. Real world noise is likely to follow a standard pattern, e.g., perhaps a target is frequently occluded and a common mistake is to place the box over the entire target rather than just the portion that is visible. This standard pattern would be much easier to learn and correct than the stocastic pattern that is applied in our experiments.

\end{document}